\documentclass{article}
\pdfoutput=1
\usepackage[final]{nips_2016} 

\usepackage[utf8]{inputenc} 
\usepackage[T1]{fontenc}    
\usepackage{url}            
\usepackage{booktabs}       
\usepackage{amsfonts}       
\usepackage{nicefrac}       
\usepackage{microtype}      
\usepackage[export]{adjustbox}

\usepackage{amssymb,amsmath}
\usepackage{algorithm,algorithmicx,algpseudocode}
\usepackage{graphicx, subfigure, caption}

\usepackage{array}

\begin{document}

\title{A K-fold Method for Baseline Estimation in Policy Gradient Algorithms}
\author
{
    Nithyanand Kota$^\dagger$, Abhishek Mishra$^\dagger$, Sunil Srinivasa$^\dagger$\\
     \normalfont{Samsung SDS Research America}\\
     $^\dagger$\normalfont{These authors contributed equally.}\\
     \texttt{\{n.kota, a2.mishra, s.srinivasa\}@samsung.com}
\And
    Xi (Peter) Chen, Pieter Abbeel \\
    \normalfont{OpenAI} \\
    \normalfont{UC Berkeley, Department of EECS} \\
    \texttt{\{peter, pieter\}@openai.com}
}


\maketitle

\begin{abstract}
    The high variance issue in unbiased policy-gradient methods such as VPG and REINFORCE is typically mitigated by adding a baseline. However, the baseline fitting itself suffers from the underfitting or the overfitting problem. In this paper, we develop a \emph{$K$-fold} method for baseline estimation in policy gradient algorithms. The parameter $K$ is the baseline estimation hyperparameter that can adjust the bias-variance trade-off in the baseline estimates. We demonstrate the usefulness of our approach via two state-of-the-art policy gradient algorithms on three MuJoCo locomotive control tasks.
\end{abstract}

\section{Introduction}
Policy gradient algorithms have garnered a lot of popularity in the area of reinforcement learning, since they directly optimize the cumulative reward function, and can be used in a straightforward manner with non-linear function approximators such as deep neural networks \citep{SchulmanTRPO,KakadeNPG}. 
They have been successfully applied towards solving continuous control tasks \citep{PetersSchaal2008,SchulmanTRPO} and learning to play Atari from raw pixels \citep{SchulmanTRPO,mnih2016asynchronous}. 
Traditional policy gradient algorithms such as REINFORCE \citep{REINFORCE} and Vanilla policy gradient \citep{Sutton99} operate by repeatedly computing estimates of the gradient of the expected reward of a policy, and then updating the policy in the direction of the gradient. 
While these algorithms provide an unbiased policy gradient estimate, the variance of the estimates is often quite high, which can severely degrade the algorithm's performance. 
Several classes of algorithms have been proposed towards reducing the variance of the policy gradient, namely actor-critic methods \citep{KondaTsisiklis2003}, using the discount factor \citep{Sutton99}, and generalized advantage estimation \citep{SchulmanGAE}.

The variance of the policy gradient can be further reduced (whilst adding no bias) by adding a baseline \citep{baseline}, which is commonly implemented as an estimator of the state-value function. 
Typically, policy gradient algorithms operate in an iterative fashion: in each iteration, they use predictions from the baseline that is fitted in the previous iteration. 
If the policy changes drastically between iterations, the baseline becomes a poor estimate of the state-value function, resulting in \emph{underfitting}. 
Alternatively, we could fit the baseline and use it to predict the value function in the same iteration. This approach suffers from the \emph{overfitting} problem. 
In the extreme case, if we only have one trajectory starting from each state, and the baseline could fit the data perfectly, then the baseline would perfectly predict the returns giving us no gradient signal at all.

In this paper, we propose a new method (which we name the $K$-fold method) for baseline estimation in policy gradient algorithms. 
The parameter $K$ is the baseline estimation hyperparameter that can adjust the bias-variance trade-off to provide a good balance between overfitting and underfitting. 
We apply our baseline prediction method in conjunction with two policy gradient algorithms -- Trust Region Policy Optimization (TRPO) \citep{SchulmanTRPO} and Truncated Natural Policy Gradient (TNPG) \citep{duan2016benchmarking}. 
We analyze the effect of different $K$ values on performance for three MuJoCo  locomotive control tasks -- Walker, Hopper and Half-Cheetah.

\section{Policy Gradient Preliminaries}
The agent's interaction with the environment is broken down into a series of $N$ episodes or trajectories. The $j^{\text{th}}$ trajectory, $j \in \{1\ldots N\}$, is notated $\tau_j=(s^j_0,a^j_0,r^j_0,s^j_1,a^j_1,r^j_1,\ldots{},s^j_{T-1},a^j_{T-1},r^j_{T-1},s^j_T)$, where $s_t, a_t$ and $r_t$ are the state, action and (instantaneous) reward respectively, as seen by the system at time $t$, and $T$ is the time horizon. The actions are chosen by the agent in each time step according to a policy $\pi$: $a_t\sim\pi(a_t|s_t)$ and the next state and reward are sampled according to the transition probability distribution $s_{t+1}, r_t \sim \mathbb{P}(s_{t+1},r_t|s_t,a_t)$.

A well-known expression \citep{REINFORCE} for the policy gradient is
\begin{equation}
  \begin{split}
    g: &=\nabla_{\theta}\mathbb{E}_{\tau}[R(\tau)] \\
    & =\mathbb{E}_{\tau}\left[\sum_{t=0}^T\nabla_{\theta}\log\pi(a_t|s_t,\theta)R(\tau)\right],
  \end{split}
\label{eqn:PGEqn1}
\end{equation}
where the policy $\pi$ is parameterized by $\theta$ and $R$ refers to the return of a trajectory under this policy.\\
An estimate of \eqref{eqn:PGEqn1} that provides better (lower-variance) policy gradients is
\begin{equation}
    \hat{g}=\frac{1}{NT}\sum_{j=0}^N\left[\sum_{t=0}^T\nabla_{\theta}\log\pi(a^j_t|s^j_t,\theta)\left(\sum_{t'=t}^T\gamma^{t'-t}r^j_{t'}-b(s^j_t)\right)\right],
\label{eqn:PGEqn2}
\end{equation}
where $N$ is the batch size, and $b(s^j_t)$ refers to the \emph{baseline}, which is an approximation of the state-value function: $b(s^j_t)\approx{}\mathbb{E}[\sum_{t'=t}^T\gamma^{t'-t}r^j_{t'}|s^j_t]$. Classically, policy gradient algorithms work in an iterative fashion as follows:
\begin{algorithm}
  \caption{Classical Policy Gradient Algorithm}
  \label{PG}
    \textbf{Initialize}: For iteration $i=0$, initialize the policy parameter $\theta_0$ randomly, and the baseline at iteration $0$, $b^{(0)}(\cdot)$ to $0$.\\
    \textbf{Iterate}: Repeat for each iteration $i, i\in 1,2,\ldots$ until convergence:
    \begin{algorithmic}[1]
      \State Sample $N$ trajectories for policy $\pi(\cdot|\cdot,\theta_{i-1})$: $\tau_{j:1,\ldots,N}$.
      \State Evaluate $\hat{g}$ as given by \eqref{eqn:PGEqn2} using predictions on the baseline at iteration $i-1$, $b^{(i-1)}(\cdot)$.
      \State Update the policy parameters using the policy gradients: $\theta_{i-1}\to\theta_{i}$.
      \State For each state $s_t$, compute the discounted returns $R_t=\sum_{t'=t}^T\gamma^{t'-t}r_{t'}$.
      \State Train (fit) a new baseline regression model $b^{(i)}(\cdot)$ with inputs $s_t$ and outputs $R_t$.
    \end{algorithmic}
\end{algorithm}

\section{$K$-Fold Method for Baseline Estimation}
\label{sec:algo}
The  policy update $\theta_{i-1}\to\theta_{i}$ above is performed in iteration $i$ using predictions from the baseline $b^{(i-1)}(\cdot)$ that was fit in the previous iteration $i-1$. An issue with this approach is \emph{underfitting}: when the policy changes a lot between iterations, the baseline predictions will be quite noisy. To circumvent this issue, we could use the data samples collected during iteration $i$ for fitting the baseline $b^{i}(\cdot)$ first, and later use the same baseline's predictions for computing the gradient and updating the policy. As noted in \citep{SchulmanGAE}, this can cause \emph{overfitting}, i.e., it reduces the variance in $\hat{g}$ at the expense of additional bias. 
Moreover, in the extreme case, if we only have one trajectory starting from each state, and if the baseline fits the data perfectly, then the gradient estimate is always zero.

In this section, we introduce the \emph{K-fold} variant of baseline prediction for policy gradient optimization (Algorithm \ref{PG}) that is more \emph{sample-efficient} than prior techniques, and helps counter both the underfitting and overfitting issues described above. 
At a high level, the algorithm operates by breaking the data samples into $K$ partitions. 
For each partition, a baseline is trained using data from all the other partitions, and the same baseline is used for predicting the value function. 
Since the baseline fitting is performed using samples from the current policy, we mitigate the problem of underfitting. 
Since we do not directly fit on the current partition's data samples, we also mitigate the overfitting issue. 
Our algorithm is general enough to be applicable to most of the algorithms for policy optimization including TRPO and TNPG.

When we divide the data into multiple partitions, it is possible to perform policy optimization via two different methods. 
In the first method, which we call the \emph{parameter-based} $K$-fold baseline estimation, we compute the gradients for each partition, use them to compute $K$ different parameters and finally average all the parameters across the  partitions to obtain the policy's new parameters. 
In the second method, which we call the \emph{gradient-based} $K$-fold baseline estimation, we compute only the gradient for each partition and use the averaged gradient to determine the policy's new parameters. 
The pseudo-code for the parameter-based and gradient-based approaches are presented in Algorithm \ref{PG:k-fold_params} and in Algorithm \ref{PG:k-fold_grads}, respectively.   

The hyperparameter $K$ essentially controls the bias-variance trade-off in the baseline estimates. On the one hand, when $K$ is small we have lesser data to fit the baseline and consequently the baseline estimates have high variance. On the other hand, when $K$ is high we get to fit the baseline with a lot of data causing the baseline estimates to have low variance, however, at an increased computational cost. The optimal value may be found using standard search techniques for hyperparameters. In this paper, we tabulate the performances of three Mujoco tasks for three values of $K$, ($K=1,2$ and $4$) in Section \ref{sec:results}.

\begin{algorithm}
  \caption{Parameter-based K-Fold Baseline Estimation for Policy Optimization}
  \label{PG:k-fold_params}
    \textbf{Initialize}: For iteration $i=0$, initialize the policy parameter $\theta_0$ randomly.\\
    \textbf{Iterate}: Repeat for each iteration $i, i\in 1,2,\ldots$ until convergence:
    \begin{algorithmic}[1]
      \State Sample $N$ trajectories from policy $\pi(\cdot|\cdot,\theta_{i-1})$: $\tau_{j:1,\ldots,N}$.
      \State For each state $s_t$ in the $N$ trajectories, compute the discounted returns $R_t=\sum_{t'=t}^T\gamma^{t'-t}r_{t'}$
      \State Partition the $N$ trajectories into $K<N$ disjoint partitions: $P_1, P_2, \ldots, P_K$, $\cup_{k=1}^K P_k = \cup_{j=1}^N\tau_j$
      \For{each partition $P_k$, $k\in\{1,\ldots,K\}$} 
        \State Train (fit) a baseline regression model $b^{(i)}_k(\cdot)$ for partition $P_k$ using the states and returns from all the remaining partitions ($P_l$, $l=1,\ldots,K$, $l\neq k$) as input and output respectively.
        \State For each partition $P_k$, initialize the policy with parameters $\theta_{i-1}$ and use any policy optimization algorithm to find optimized policy parameters $\theta_{i}^k$: $\theta_{i-1}\to\theta_{i}^k$.
      \EndFor{}
      \State Update the policy parameters $\theta_{i-1}\to\theta_i$ as the average of all the optimized policy parameters obtained, i.e., $\theta_i = \frac{1}{K}\sum_{k=1}^K \theta_i^k$.
    \end{algorithmic}
\end{algorithm}

\begin{algorithm}
  \caption{Gradient-based K-Fold Baseline Estimation for Policy Optimization}
  \label{PG:k-fold_grads}
    \textbf{Initialize}: For iteration $i=0$, initialize the policy parameter $\theta_0$ randomly.\\
    \textbf{Iterate}: Repeat for each iteration $i, i\in 1,2,\ldots$ until convergence:
    \begin{algorithmic}[1]
      \State Sample $N$ trajectories from policy $\pi(\cdot|\cdot,\theta_{i-1})$: $\tau_{j:1,\ldots,N}$.
      \State For each state $s_t$ in the $N$ trajectories, compute the discounted returns $R_t=\sum_{t'=t}^T\gamma^{t'-t}r_{t'}$
      \State Partition the $N$ trajectories into $K<N$ disjoint partitions: $P_1, P_2, \ldots, P_K$, $\cup_{k=1}^K P_k = \cup_{j=1}^N\tau_j$ 
      \For{each partition $P_k$, $k \in \{1,\ldots,K\}$} 
        \State Train (fit) a baseline regression model $b^{(i)}_k(\cdot)$ for partition $P_k$ using the states and returns from all the remaining partitions ($P_l$, $l=1,\ldots,K$, $l\neq k$) as input and output respectively.
        \State Evaluate the gradient for partition $P_k$, $g_k$ using predictions from the baseline $b^{(i)}_k(\cdot)$.
      \EndFor{}
      \State Compute the average gradient $g=\frac{1}{K}\sum_{k=1}^Kg_k$.
      \State Use the average gradient $g$ in any  policy optimization algorithm to update the policy parameters $\theta_{i-1}\to\theta_i$.
    \end{algorithmic}
\end{algorithm}

\section{Tasks}
\label{sec:task}
We evaluate our algorithm on three MuJoCo  locomotive tasks - Hopper, Walker and Half-Cheetah using the rllab \citep{duan2016benchmarking} platform. These tasks are chosen since they are complex, yet allow for fast experimentation.

\textbf{Hopper} is a planar monopod robot with $4$ rigid links and $3$ actuated joint. The aim is to learn a stable hopping gait without falling, and avoid local minimum gaits like diving forward. The observation space is $20$-dimensional including joint angles, joint velocities, center of mass coordinates and the constraint forces. The reward function is given by $r(s,a):= v_{x}-0.005*||a||_2^2+1$. The episode is terminated when $z_{body}<0.7$ or $|\theta_y|<0.2$ where  $z_{body}$ is the z-coordinate of the center of mass, and $\theta_y$ is the forward pitch of the body.

\textbf{Walker} is a planar bipedal robot with $7$ rigid links and $6$ actuated joints. The observation space is $21$-dimensional including joint angles, joint velocities and center of mass coordinates.The reward function is given by $r(s,a):= v_{x}-0.005*||a||_2^2$. The episode is terminated when $z_{body}<0.8$ or $z_{body}>2$, or when   $|\theta_y|>1$ where  $z_{body}$ is the z-coordinate of the center of mass, and $\theta_y$ is the forward pitch of the body.

\textbf{Cheetah} is a planar bipedal robot with $9$ links and $6$ actuated joints. The observation space is $20$-dimensional including joint angles, joint velocities and center of mass coordinates.The reward function is given by $r(s,a):= v_{x}-0.05*||a||_2^2$.

\section{Experimental Setup}
The experimental setup used to generate the results is described below.

\textbf{Performance Metrics:} For each experiment (running a specific algorithm on a specific task), we define its performance as $\frac{1}{I}\sum_{i=1}^IR_i$, 
where $I$ is the number of training iterations and $R_{i}$ the undiscounted average return for the $i$th iteration. This is essentially the area under the average return curve. We use $5$ random starting seeds and report the performance averaged over all the seeds.

\textbf{Policy Network:} We employ a feed-forward Multi-Layer Perceptron (MLP) with $3$ hidden layers of sizes $100$, $50$ and $25$ with tanh nonlinearities after the first two hidden layers that maps states to the mean of a Gaussian distribution. We used the conjugate gradient method for policy optimization.

\textbf{Baseline:} We use a Gaussian MLP for the baseline representation as well, with $2$ hidden layers of size $32$ each (and a tanh nonlinearlity after the first hidden layer). The baseline is fitted using the ADAM first order optimization method \citep{kingma2014adam}. 
In each iteration, we utilized $10$ ADAM steps, each with a batch size of $50$.

\textbf{Hyperparameters:} For all the experiments, we use the same experimental setup described in \citep{duan2016benchmarking}, and listed in Table \ref{table:exp_setup}.

\begin{table}[hbt]
  \centering
  \caption{\label{table:exp_setup}Basic experimental setup parameters.}
  \begin{tabular}{|c|c|}
    \hline
    Discount Factor & $0.99$ \\
    Horizon & $500$ \\
    Number of Iterations & $500$ \\ 
    Step size for both TRPO and TNPG & 0.08\\
    \hline
  \end{tabular}
\end{table}

\section{Results}
\label{sec:results}
We evaluate both the parameter-based and the gradient-based K-fold algorithms on the three locomotive tasks listed above and on two policy gradient algorithms, TRPO and TNPG. In the tables below, we provide the $mean(returns)$ and the $std(returns)$ numbers. We also highlight (boldface) the cases where the $mean-std$ numbers were the best.

\subsection{Results with TRPO, Data size = 50000}
First, we evaluate the $K$-fold methods (parameter-based and gradient-based) with the Trust Region Policy Optimization (TRPO) algorithm. We used a data (sample) size of $50,000$. The results are tabulated in Table \ref{table:TRPO_50000}
\begin{table}[hbt]
  \centering
  \caption{\label{table:TRPO_50000}Results with TRPO and data size of $50,000$.}
  \begin{tabular}{|c|c|c|c|}
    \hline
    \multicolumn{4}{|c|}{\textbf{Parameter-based}} \\
    \hline
    Task & $K=1$ & $K=2$ & $K=4$\\
    \hline
    Walker & $911.0\pm681.0$ & $\mathbf{1015.7\pm327.3}$ & $938.7\pm462.1$ \\
    \hline
    Hopper & $727.7\pm242.6$ & $723.7\pm190.5$ & $\mathbf{721.4\pm149.5}$\\
    \hline
    Cheetah & $\mathbf{1595.1\pm404.4}$ & $1528.5\pm406.6$ & $1383.8\pm356.1$\\
    \hline
  \end{tabular}
\end{table}

\begin{figure}[h]
    \centering
    \begin{subfigure}{}
        \centering
        \includegraphics[scale=0.22]{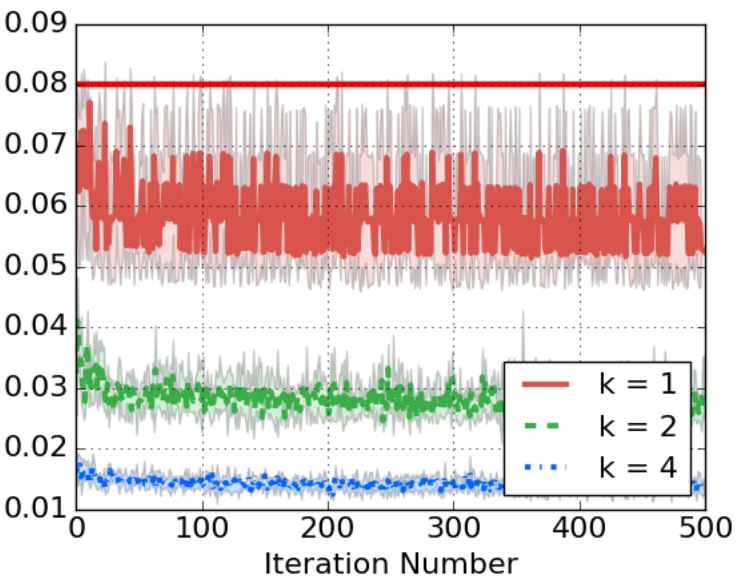}
    \end{subfigure}
    \begin{subfigure}{}
        \centering
        \includegraphics[scale=0.22]{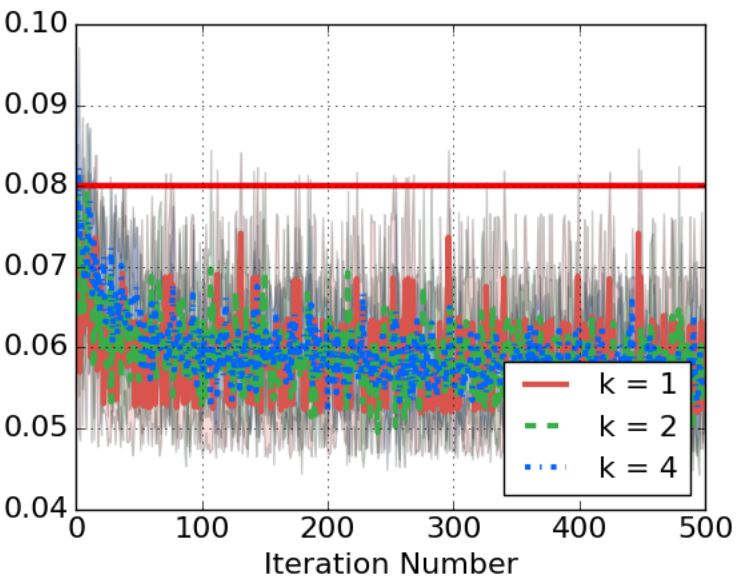}
    \end{subfigure}    
    \begin{subfigure}{}
        \centering
        \includegraphics[scale=0.22]{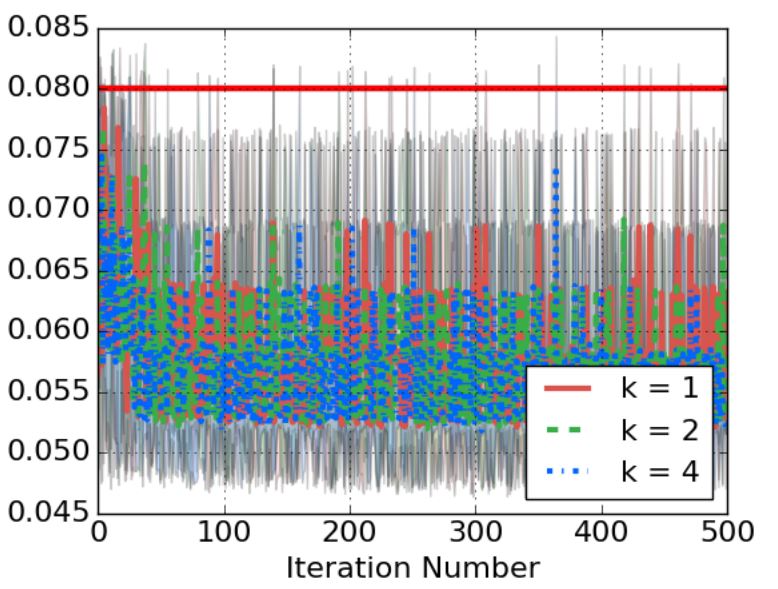}
    \end{subfigure}
    \caption{Mean $\pm$ std KL divergence numbers for $K=1, 2$ and $4$ for the Cheetah task. The solid red line depicts the mean KL divergence constraint of $0.08$. (Left): the KL divergence numbers for the parameter-based method - they are seen to scale down linearly with increasing $K$. (Center): the KL divergence numbers for the parameter-based method with scaled step sizes and (Right): the KL divergence numbers for the gradient-based method. For the latter two methods, KL divergence numbers are very comparable across the three values of $K$.}
    \label{figure:mean_KL}
\end{figure}

Under the parameter-based approach it is observed that for $K>1$, the mean KL divergence between consecutive policies is much lower than the prescribed  constraint value of $0.08$. In fact, it is seen (see Figure \ref{figure:mean_KL} (Left)) that the KL divergence values scales down linearly with increasing values of $K$. A lower KL divergence between consecutive policies implies reduced exploration for larger $K$. To overcome this shortcoming, we perform an additional experiment for the parameter-based method where the step-size was scaled up by a factor of $K$. After scaling up the step sizes, we observe that the mean KL divergence values are comparable across all values of $K$ (see Figure \ref{figure:mean_KL} (Center)). Table \ref{table:TRPO_50000_scaled} provides the return numbers for the parameter-based approach with scaled step sizes.

\begin{table}[hbt]
  \centering
  \caption{\label{table:TRPO_50000_scaled}Results with TRPO with scaled step-sizes and data size of $50,000$.}
  \begin{tabular}{|c|c|c|c|}
    \hline
    \multicolumn{4}{|c|}{\textbf{Parameter-based (with scaled step sizes)}} \\
    \hline
    Task & $K=1$ & $K=2$ & $K=4$\\
    \hline
    Walker & $911.0\pm681.0$ & $1053.7\pm427.1$ & $\mathbf{1188.8\pm251.8}$\\
    \hline
    Hopper & $727.7\pm242.6$ & $\mathbf{811.3\pm112.7}$ & $745.1\pm114.1$\\
    \hline
    Cheetah & $1595.1\pm404.4$ & $1496.3\pm401.4$ & $\mathbf{1600.0\pm237.9}$\\
    \hline
  \end{tabular}  
\end{table}

The scaling idea above was an ad-hoc approach for improving the exploration for cases with $K>1$. Alternatively, we can use the gradient-based method, wherein averaging is performed across the $K$ parameter gradients themselves, naturally ensuring that the mean KL divergence values remain similar and below the constraint (see Figure. \ref{figure:mean_KL} (Right)). Table \ref{table:TRPO_50000_gradient} lists the return numbers for the gradient-based $K$-fold variant. Figure \ref{figure:AvgReturn} (left) plots the average return mean and std numbers for the Hopper Task.

\begin{table}[hbt]
  \centering
  \caption{\label{table:TRPO_50000_gradient}Results with TRPO with the gradient-based approach and a data size of $50,000$.}
  \begin{tabular}{|c|c|c|c|}
    \hline
    \multicolumn{4}{|c|}{\textbf{Gradient-based}} \\
    \hline
    Task & $K=1$ & $K=2$ & $K=4$\\
    \hline
    Walker & $911.0\pm681.0$ & $1035.0\pm491.1$ & $\mathbf{1092.8\pm401.2}$\\
    \hline
    Hopper & $727.7\pm242.6$ & $\mathbf{786.0\pm171.1}$ & $847.7\pm274.0$\\
    \hline
    Cheetah & $1595.1\pm404.4$ & $1664.1\pm337.1$ & $\mathbf{1676.1\pm333.4}$\\
    \hline
  \end{tabular}  
\end{table}

\subsection{Results with TNPG, Data size = 50000}
Next, we evaluate the $K$-fold gradient-based method on another policy gradient algorithm, the Truncated Natural Policy Gradient (TNPG). The results are tabulated in \ref{table:TNPG_50000_gradient}.
\begin{table}[hbt]
  \centering
  \caption{\label{table:TNPG_50000_gradient}Results with TNPG with the gradient-based approach and data size of $50,000$.}
  \begin{tabular}{|c|c|c|c|}
    \hline
    \multicolumn{4}{|c|}{\textbf{Gradient-based}} \\
    \hline
    Task & $K=1$ & $K=2$ & $K=4$\\
    \hline
    Walker & $863.1\pm714.7$ & $\mathbf{930.3\pm597.2}$ & $984.2\pm674.7$\\
    \hline
    Hopper & $895.4\pm173.4$ & $\mathbf{858.3\pm93.7}$ & $775.5\pm246.0$\\
    \hline
    Cheetah & $1635.6\pm352.9$ & $1602.3\pm392.3$ & $\mathbf{1627.8\pm314.5}$\\
    \hline
  \end{tabular}  
\end{table}

\subsection{Results with TRPO and TNPG, Data size = 5000}
\begin{figure}[hbt]
    \centering
    \begin{subfigure}{}
        \centering
        \includegraphics[scale=0.3]{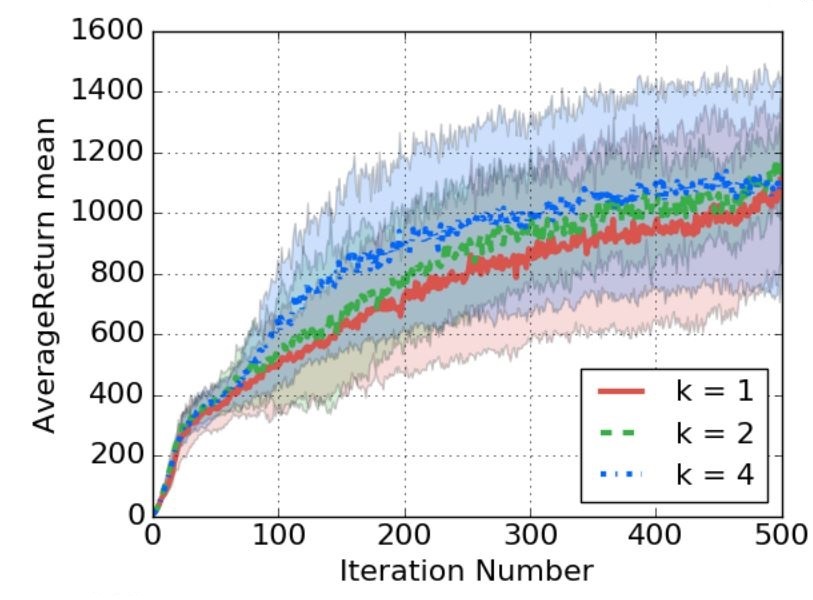}
    \end{subfigure}
    \begin{subfigure}{}
        \centering
        \includegraphics[scale=0.3]{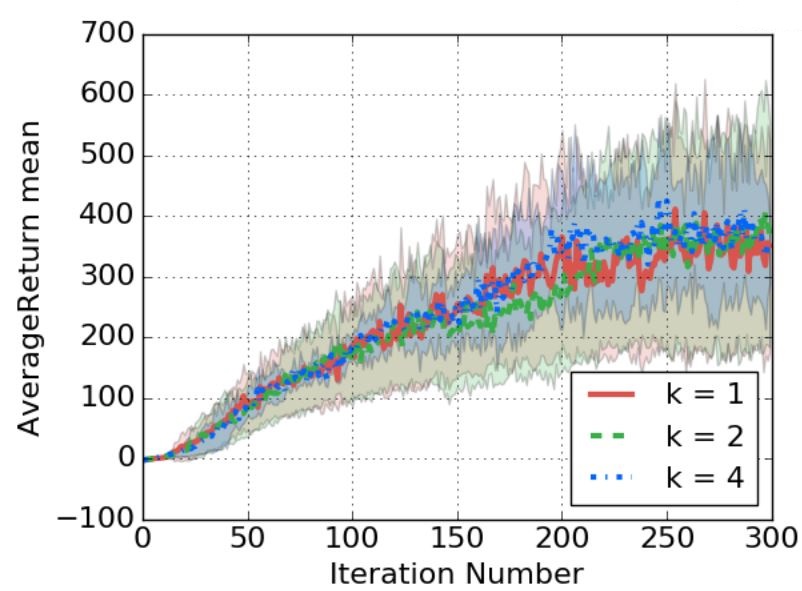}
    \end{subfigure}
    \caption{Average mean $\pm$ standard deviation trajectories. (Left): Hopper task with TRPO and a data size of $50000$. (Right): Walker2D task with TNPG and a data size of $5000$.}
    \label{figure:AvgReturn}
\end{figure}

Finally, we evaluate the $K$-fold gradient-based method on lower data size. The results are presented in Table \ref{table:TRPO_TNPG_5000_gradient}. Figure \ref{figure:AvgReturn} (Right) plots the average return mean and std numbers for the Walker Task.

\begin{table}[hbt]
  \centering
  \caption{\label{table:TRPO_TNPG_5000_gradient}Results with TRPO and TNPG with the gradient-based approach and data size of $5,000$.}
  \begin{tabular}{|c|c|c|c|}
    \hline
    \multicolumn{4}{|c|}{\textbf{TRPO}} \\
    \hline
    Task & $K=1$ & $K=2$ & $K=4$\\
    \hline
    Walker & $\mathbf{375.1\pm170.2}$ & $292.8\pm167.9$ & $389.6\pm225.1$\\
    \hline
    Hopper & $\mathbf{346.9\pm36.2}$ & $348.3\pm41.5$ & $319.5\pm13.4$\\
    \hline
    Cheetah & $666.5\pm364.0$ & $\mathbf{727.0\pm313.2}$ & $591.5\pm184.4$\\
    \hline
  \end{tabular}
  
  \begin{tabular}{|c|c|c|c|}
    \hline
    \multicolumn{4}{|c|}{\textbf{TNPG}} \\
    \hline
    Task & $K=1$ & $K=2$ & $K=4$\\
    \hline
    Walker & $299.4\pm154.0$ & $316.6\pm164.6$ & $\mathbf{336.7\pm91.9}$\\
    \hline
    Hopper & $331.4\pm42.6$ & $317.3\pm29.5$ & $\mathbf{344.7\pm31.9}$\\
    \hline
    Cheetah & $\mathbf{609.5\pm215.3}$ & $445.9\pm228.8$ & $445.9\pm181.9$\\
    \hline
  \end{tabular}   
\end{table}

\section{Summary}
In this paper, we proposed a new method for baseline estimation in policy gradient algorithms. We tested our method across three environments: Walker, Hopper, and Half-Cheetah; two policy gradient algorithms: TRPO and TNPG; two data sizes: $50000$ and $5000$; and three different values of $K$: $1$, $2,$ and $4$. We find that hyperparameter $K$ is a useful parameter to adjust the bias-variance trade-off, in order to achieve a good balance between overfitting and underfitting. While no single value of $K$ was found to result in the best average return across all the cases, it was observed that values other than $K=1$ can provide improved returns in certain cases, thus highlighting the need for a hyperparameter search across the candidate $K$ values. As a part of future work it will be interesting to study the benefit that the $K$-fold method provides for other environments, policy gradient algorithms, step sizes and batch sizes.

\section*{Acknowledgement}
The authors would like to thank Girish Kathalagiri, Aleksander Beloi, Luis Carlos Quintela and Atul Varshneya for their invaluable help and suggestions during various stages of this project.
\medskip

\bibliographystyle{unsrt}


\end{document}